\documentclass{article}

\usepackage{arxiv}

\usepackage[utf8]{inputenc} 
\usepackage[T1]{fontenc}    
\usepackage{hyperref}       
\usepackage{url}            
\usepackage{booktabs}       
\usepackage{amsfonts}       
\usepackage{nicefrac}       
\usepackage{microtype}      
\usepackage{lipsum}
\usepackage{graphicx}

\usepackage{amsmath}
\usepackage{amssymb}
\usepackage{breqn}
\usepackage{dsfont}

\usepackage{setspace}

\title{Is Discriminator a Good Feature Extractor?}

\author{
 Xin Mao \\
  Hong Kong University of Science and Technology\\
  HongKong, China \\
  \texttt{xmaoac@connect.ust.hk} \\
   \And
 Zhaoyu Su \\
  Hong Kong University of Science and Technology\\
  HongKong, China \\
  \texttt{zsuad@connect.ust.hk} \\
  \And
 Pin Siang Tan \\
  Hong Kong University of Science and Technology\\
  HongKong, China \\
  \texttt{pstan@ust.hk} \\
  \And
 Jun Kang Chow \\
  Hong Kong University of Science and Technology\\
  HongKong, China \\
  \texttt{jkchow@connect.ust.hk} \\
  \And
 Yu-Hsing Wang \\
  Hong Kong University of Science and Technology\\
  HongKong, China \\
  \texttt{ceyhwang@ust.hk} \\
}

\begin{document}
\maketitle

\begin{abstract}
  The discriminator from generative adversarial nets (GAN) has been used by some researchers as a feature extractor in transfer learning and appeared worked well. However, there are also some studies that believe this is the wrong research direction because intuitively the task of the discriminator focuses on separating the real samples from the generated ones, making features extracted in this way useless for most of the downstream tasks. To avoid this dilemma, we first conducted a thorough theoretical analysis of the relationship between the discriminator task and the characteristics of the features extracted. We found that the connection between the task of the discriminator and the feature is not as strong as was thought, for that the main factor restricting the feature learned by the discriminator is not the task of the discriminator itself, but the need to prevent the entire GAN model from mode collapse during the training. From this perspective and combined with further analyses, we found that to avoid mode collapse in the training process of GAN, the features extracted by the discriminator are not guided to be different for the real samples, but divergence without noise is indeed allowed and occupies a large proportion of the feature space. This makes the features learned more robust and helps answer the question as to why the discriminator can succeed as a feature extractor in related research. Consequently, to expose the essence of the discriminator extractor as different from other extractors, we analyze the counterpart of the discriminator extractor, the classifier extractor that assigns the target samples to different categories. We found the performance of the discriminator extractor may be inferior to the classifier based extractor when the source classification task is similar to the target task, which is the common case, but the ability to avoid noise prevents the discriminator from being replaced by the classifier. Last but not least, as our research also revealed a ratio playing an important role in GAN's training to prevent mode collapse, it contributes to the basic GAN study.
\end{abstract}


\section{Introduction}

A feature is a compressive representation of data with small information loss for potential target tasks. A feature extractor is an important part of transfer learning, which describes the transferring of knowledge from some data domains and learning tasks to other domains and tasks \cite{pan_survey_2009}. An example of feature extractor based transfer learning is that a ResNet \cite{he_deep_2016} based image classifier can be first trained on ImageNet dataset \cite{deng_imagenet:_2009} to get a feature extractor, and then be fine-tuned on the CIFAR dataset \cite{krizhevsky_learning_2009} to reduce the efforts of model re-calibration.  Another example is that an autoencoder \cite{baldi_autoencoders_2012}  can be trained by an encoding-decoding task and then provides its encoder as the feature extractor for the downstream tasks. Feature extractor works because, while some features are task-specific, there are also some features can be shared between tasks \cite{liu_adversarial_2017,pan_survey_2009}, and the features extracted are compressed and are more linearly separable \cite{pan_survey_2009}, which are easier to be handled. The use of feature extractor in transfer learning has the ability to help to handle the lack of labeled data \cite{cui_large_2018}, prevent overfitting \cite{gao_deep_2018} and promote metalearning \cite{vilalta_perspective_2002}, etc.

\begin{figure}[t]
   \begin{center}
      \includegraphics[width=0.8\linewidth]{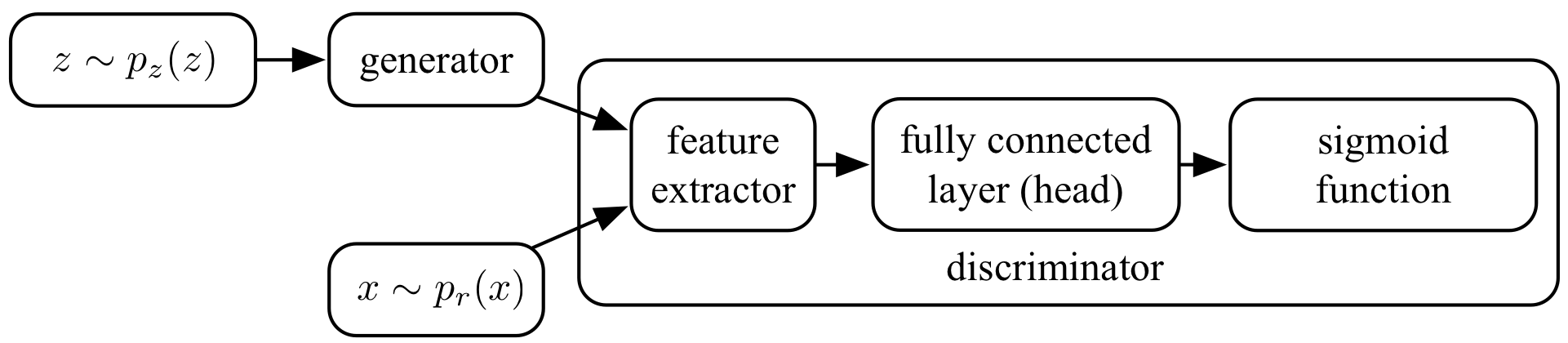}
   \end{center}
      \caption{Generative adversarial nets structure. $p_z(z)$ is the probability density function of the random vector $z$, and $p_r(x)$ is the density function of the real data domain.}
   \label{fig1}
\end{figure}

When the source task is unsupervised, in addition to the autoencoder that has been mentioned, Generative Adversarial Nets (GAN) \cite{goodfellow_generative_2014} can also be chosen as the source learning framework to provide feature extractors. GAN was originally developed to generate data with the same distribution as the real dataset. For example, if there are some handwritten words, it can learn to generate words that cannot be distinguished from the real ones by human eyes. GAN’s structure is shown in Figure \ref{fig1}. It consists of a generator and a discriminator, where the generator receives a random vector as the input and outputs a fake sample. The discriminator then receives a sample and judges whether it is from the real data domains or is generated. GAN’s objective function is shown in Equation (\ref{eq1}). From this, the two sides play a mini-max game, whereby the generator tries to fool the discriminator to make it unable to distinguish the fake samples from the real ones, while the discriminator has the opposite goal. When a Nash Equilibrium \cite{nash_equilibrium_1950} is achieved, where neither the generator nor the discriminator can improve its performance by just changing itself, the fake sample will be like the real one, and the training will be completed.

To provide the feature extractor, the GAN can either use its generator or its discriminator. The extractor based on the generator was proposed by Jeff et al. \cite{donahue_adversarial_2016}, together with the BiGANs model. This model adds an auxiliary neural network mapping the generated sample to the corresponding random vector, which is the input of the generator. Once the GAN's training is completed, this auxiliary neural network can be used as the feature extractor, which accepts a real data sample and outputs a vector as its feature representation. This works because the change of random vector causes semantic variations of the generated samples, and playing as a feature it is well disentangled \cite{donahue_adversarial_2016}. To improve the performance of this kind of extractor, one can use InfoGAN \cite{chen_infogan:_2016} to replace the GAN part in the BiGANs. This utilizes the mutual information to enhance the random vectors to make them more disentangled. However, this kind of feature extractor has its own particular defect, in that one cannot ensure that the mapping from the random vector to the sample by the generator is invertible \cite{odena_conditional_2017}. So even if the random vector is a good feature representation and is disentangled, the learned feature could be bad, for it is not the aforementioned random vector itself. What's more, it is reported that disentanglement does not always lead to good feature representation \cite{locatello_challenging_2018}, which makes things worse.

The feature extractor based on discriminator was introduced by Radford et al. \cite{radford_unsupervised_2015} together with the DCGAN model. In this work, after the training of the DCGAN on some datasets, the author used the convolutional features of the discriminator as the base for the downstream classification tasks on the CIFAR-10 and SVHN datasets and achieved a good performance. After this work, the discriminator based feature extractor was tried by some researchers in different fields to help them achieve their goals \cite{lin_marta_2017,zhang_unsupervised_2018}. However, despite these successes, there are some doubts -- as pointed by Springenberg \cite{springenberg_unsupervised_2016}, intuitively, as the target of the discriminator is to distinguish the generated samples from the real samples, it will just focus on the difference between these two kinds of samples. However, what makes sense is the difference between the real samples, which are the samples used by the downstream tasks. These doubts led us to check whether the successes of the discriminator based feature extractor was by chance, or supported by some mechanisms that were not noticed. Until we answer this question, we cannot be sure whether research on the discriminator extractor has potential.

If the success of the discriminator extractor is supported, there are three possibilities. a) Although the task of the discriminator does not distinguish different real samples, the features extracted for those samples could be guided to be different by other means; b) Even if there is no such guidance, the features extracted from those samples are allowed to be different in a meaningful way; c) The specific task only restricts the final state of the discriminator, but before the training is completed, there may be something to cause the features of the real samples to be different. To check these possibilities, in this research we use mathematical tools to analyze the relationship between the task of the discriminator and its features extracted, and assess the training process of GAN to find the characteristics of features extracted by the discriminator in each step. Besides this, so as to understand a particular phenomenon we must also understand its counterpart, we analyze the classifier based feature extractor using a similar approach, because unlike discrimination, the classification task focuses on the difference between the target samples. In addition to theoretical analyses, we also conducted an experimental comparison between the discriminator and classifier based feature extractors to support our findings.

Our research makes the following contributions: a) It reveals that the main factor restricting the feature learned by the discriminator is not the task of the discriminator itself, but the need to prevent the entire GAN model from mode collapse during the training. b) It reveals that discriminator will not guide the features extracted to be different but allows such difference with no noise interference. This difference is proven to influence a large proportion of the feature space and helps explain why the discriminator can be used as a successful feature extractor in certain cases. c) It reveals that the classifier forces the features to be different, based on the categories to be classified, so in cases where the source classification task is similar to the target task, it is a better feature extractor than discriminator. However, the classifier cannot force features to get rid of the influence from noise, so it cannot replace the discriminator. d) It reveals a special ratio is playing an important role in mode collapse, which has the theoretical and practical value in basic GAN study.

\section{Analyze discriminator as feature extractor}

To analyze the characteristics of discriminator as feature extractor, we start from the objective function of GAN:

\begin{dmath}
    \min_G \max_D V(D, G) = E_{x \sim p_r(x)} \log D(x) + E_{z \sim p_z(z)} \log \{1 - D[G(z)]\}
    \label{eq1}
\end{dmath}

In this equation, $G$ represents the generator; $D$ represents the discriminator; $p_z(z)$ is the probability density function of the random vector; $p_r(x)$ is the density function of the real distribution; $G(z)$ is the generated sample; and $D(x)$ is the probability that a sample is from the real data distribution. This equation can be rewritten as:

\begin{dmath}
    \min_G \max_D V(D, G)= \int_x \{p_r(x) \log D(x) + p_g(x) \log [1 - D(x)]\} dx
    \label{eq2}
\end{dmath}

\noindent where the new term $p_g$ is the density of the generated distribution. This equation leads to the following proposition

\textbf{Proposition 1} \cite{goodfellow_generative_2014}\textbf{.} when $G$ or $p_g(x)$ is fixed, the optimal discriminator $D^{*}$ is

\begin{equation}
    D^{*} (x) = \frac{p_r(x)}{p_r(x) + p_g(x)}
    \label{eq3}
\end{equation}

\textit{Proof:} $V(D, G)$ will get its minimum when $f(x) = p_r(x) \log D(x) + p_g(x) \log[1 - D(x)]$ is minimum. For this situation, when $G$ is fixed, the following equation holds

\begin{equation}
    \frac{\partial f}{\partial D} = \frac{p_r(x)}{D(x)} + \frac{p_g(x)}{D(x) - 1} = 0
    \label{eq4}
\end{equation}

\noindent which means $D^{*}(x) = \frac{p_r(x)}{p_r(x)+p_g(x)}$.

Introduce the Definition

\textbf{Definition 1.} The support of a distribution $p$, which is $Supp(p)$, is the set of samples $x$ where the density $p(x) \neq 0$.

Combining this with Proposition 1, we have

\textbf{Corollary 1.} $D^{*}(x)$ is controlled by the ratio $\alpha(x)=\frac{p_g(x)}{p_r(x)}$ for $x \in Supp(p_r)$. That is $D^{*}(x)=\frac{1}{\alpha(x) + 1}$.

Ruled by this corollary, together with that the real data density $p_r(x)$ is given, the generated sample distribution is just constrained by $p_g(x)$, which is determined by the optimal state of the generator. This optimal state is coordinated by the following Proposition.

\textbf{Proposition 2.} When $D$ is given and set to be $D^{*}$, the optimal $p_g(x)$ for $x \in Supp(p_r)$ is only constrained by the ratio $\alpha_0(x)=\frac{p_{g_0(x)}}{p_r(x)}$, where $p_{g_0}$ is the fixed generated distribution density in the previous step to guide $D$ to be $D^{*}$.

\textit{Proof:} in Equation (\ref{eq2}), for $x \in Supp(p_r) \bigcup Supp(p_g)$, when $D^{*}$ is fixed, only

\begin{dmath}
    C(G) = \int_x p_g(x) \log[1 - D^{*}(x)] dx = \int_x p_g(x) \log \frac{p_{g_0}(x)}{p_{g_0}(x) + p_r(x)} dx
    \label{eq5}
\end{dmath}

\noindent corresponds to minimize $V$, so that this is the objective function for $G$. Because of that $\log \frac{p_{g_0}(x)}{p_{g_0}(x) + p_r(x)}=0$ when $x \notin Supp(p_r)$, this equation can be simplified as $\int_{x \sim  Supp(p_r)} p_g(x) \log \frac{\alpha_0(x)}{\alpha_0(x) + 1}dx$. As $\log\frac{\alpha_0(x)}{\alpha_0(x)+1}$ is frozen, this rewritten equation is a linear optimization function with the constraints that $p_g(x) \geq 0$ and $\int_{x \sim Supp(p_r)} p_g(x)dx \le 1$. In this situation, the optimal $G$, representing by $p_g$, will just be controlled by $\alpha_0$ and the constraints.

Further, the $\alpha$ mentioned above is controlled by the following Corollary.

\textbf{Corollary 2.} If the generator and the discriminator are trained alternately and achieve optimal at each step, $\alpha(x)=\frac{p_g(x)}{p_r(x)}$ should be invariant to $x \in Supp(p_r)$ and only change based on steps, or the GAN will fall into mode collapse.

\textit{Proof:} model collapse describes that the generator just learns to produce a limited range of samples from the real distribution \cite{barnett_convergence_2018}. As described, $G$'s objective function (Equation \ref{eq5}) is a linear function with linear constraints. The linearity will cause the optimal results $p_g(x)\ \rightarrow 0$ for all $x \notin \arg \min_x \alpha(x)$. If $\alpha(x)$ is variant to $x \in Supp(p_r)$, then $\exists x^' \in Supp(p_r)$ causes $\alpha(x^') \neq \min \alpha(x)$, and $p^*(x^') \rightarrow 0$. Once this happens, in the next training iteration. As $\alpha(x) = \frac{p_g(x)}{p_r(x)}$, the $x$ inside $Supp(p_r)$ whose $p_g(x) \rightarrow 0$ before will now be pushed to have a high generated probability density, while the rest inside the support will be pushed to have $p_g \rightarrow 0$. Therefore, the distribution of generated sample will jump from a limited range to another limited range; the model collapse appears; and the train of GAN will fail.

Mode collapse should be avoided to make the GAN training successful, so in this case, $\alpha(x)$ must be a constant for $x \in Supp(p_r)$. Combined with Corollary 1 that the optimal state of discriminator is just controlled by the $\alpha(x)$, the following Corollary will reveal the strict constraint the optimal state of discriminator must obey.

\textbf{Corollary 3.} For a successfully trained GAN, in any stage of its training, $D^*(x)$ should be invariant to $x \in Supp(p_r)$.

This property of the discriminator has been revealed for a special case in previous studies, and it is described using the following Proposition.

\textbf{Proposition 3} \cite{goodfellow_generative_2014}\textbf{.} The Nash Equilibrium of GAN is achieved if and only if $p_g(x)=p_r(x)$. In this situation, $D^{*}(x)=\frac{1}{2}$ for $x \in Supp(p_r) \cup Supp(p_g)$.

\textit{Proof:} When the Nash Equilibrium archives, $D(x)$ will be the $D^*(x)$ in Proposition 1. The objective function in Equation (\ref{eq2}) can be rewritten as

\begin{dmath}
    C(G)=\int_{x} [p_r(x) \log \frac{p_r(x)}{p_r(x)+p_g(x)} + p_g(x) \log \frac{p_g(x)}{p_r(x)+p_g(x)} ]dx
    \label{eq6}
\end{dmath}

\noindent which can be further rewritten as

\begin{dmath}
    C(G) = -\log(4) + D_{KL}(p_r \Vert \frac{p_r + p_g}{2}) + D_{KL}(p_g \Vert \frac{p_r + p_g}{2})
    \label{eq7}
\end{dmath}

This equation will achieve minimum if and only if $p_r(x)=p_g(x)$,  where $D^*(x)=\frac{p_r(x)}{p_r(x) + p_g(x)} = \frac{1}{2}$.

This proposition implies that when a GAN is trained completely, $D^*(x)$ will be $\frac{1}{2}$, which is invariant to $x \in Supp(p_r)$ as $Supp(p_r)=Supp(p_g)$. Our proof extends this conclusion by letting the outputs of $D^*(x)$ in every training stage to be invariant to $x \in Supp(p_r)$. And our proof also extends the mechanisms behind this invariance, that for the final state, the invariance may be due to the essence of the discrimination task, but for the entire training process, the invariance is caused by the need to prevent the entire GAN model from mode collapse. This shows that there is no difference between the final training step and the earlier steps.

Utilizing this invariance, we can evaluate the characteristics of the features extracted by the discriminator. To do so, we should first separate the component of the discriminator acting as the feature extractor from the others. In most cases, the discriminator $D$ is a neuron network consisting of three components. The first component is the feature extractor $f$. The second component is a fully connected layer without bias, which works as a matrix $A$ applying a linear transform to the output of the first component. The third component is a sigmoid function $\sigma$ mapping the output from the second component to the range $(0, 1)$, which can be thought of as the probability that a sample is real. Based on this decomposition, the discriminator can be represented by Equation (\ref{eq8}), while Equation (\ref{eq9}) represents the third component, the sigmoid function.

\begin{dmath}
    D(x)=\sigma[A f(x)]
    \label{eq8}
\end{dmath}

\begin{dmath}
    \sigma(y)=\frac{1}{1 + e^{-y}}
    \label{eq9}
\end{dmath}

As shown, the sigmoid function $\sigma$ is a monotone function. Together with that $D(x)$ should be invariant to $x \in Supp(p_r)$ as shown in Corollary 3, the output of the second component $b = A f(x)$ should be invariant for $x \in Supp(p_r)$. As the output is a linear transformation of the features from the extractor, it should be ruled by the following propositions.

\textbf{Proposition 4} \cite{strang_introduction_1993}\textbf{.} For a matrix $A$, its nullspace $N(A)$ is the orthogonal complement of its row space $C(A^T)$. For every $y$ in $b = Ay$, it can be uniquely split into a row space component $y_r$ and a nullspace component $y_n$, where $y = y_r + y_n$.

\textbf{Proposition 5} \cite{strang_introduction_1993}\textbf{.} Every vector $b$ in the column space comes from one and only one vector in the row space.

Applying these to our features extracted, we have the following Corollary

\textbf{Corollary 4.} For a successfully trained GAN, in any stage of its training, the features $y = f(x)$ extracted by the discriminator for the real data can be expressed as $y(x)= c + y_n(x)$, where $y_n$ is inside the nullspace of the matrix $A$, the second component of the discriminator, and $c$ is a vector invariant to $x$, which is inside the row space of $A$.

\textit{Proof:} As shown above, $b = Af(x)$ is invariant to $x \in Supp(p_r)$, which means $b$ is a vector inside $A$'s column space invariant to $x$. Supposing $f(x) = f_r(x) + f_n(x)$, where $f_r(x)$ is inside $A$'s row space and $f_n(x)$ is inside $A$'s nullspace (Proposition 4), then $f_r(x)$ is a vector invariant to $x$ (Proposition 5).

This leads to the following Theorem

\textbf{Theorem 1.} For a successfully trained GAN, in any stage of its training, the features extracted by the discriminator are not guaranteed to be diverse in the support of the real data distribution. However, such divergence is allowed and will influence a large proportion of the feature space, and these features will get rid of the influence of noise and be more robust.

\textit{Proof:} From Corollary 4, the features extracted by discriminator can be decomposed into $f(x) = c +f_n(x)$. The divergence of $f(x)$ is represented by the divergence of $f_n(x)$. The latter is inside the nullspace of the matrix $A$ as the second component of the discriminator. The matrix $A$ provides no guide to make $f_n(x)$ variant for $x \in Supp(p_r)$, but as the null space is the only restriction, the difference in features of the real samples is indeed allowed. As the null space is far larger than the row space, the mentioned difference influences a large part of the feature, and these differences can be explained by the differences in different samples. Finally, as the null space is orthogonal to the row space, the features explaining the difference between the real samples and the generated samples, which in most cases are the noise, will be excluded from the null space. As all real samples have a similar row space features, the entire features for the real samples will eliminate the influence of the noise.

Considering that in most cases the target domain will be the same as or very close to the real domain, the conclusion relating to the real samples could be transfered to the target domain. This helps check the possibilities \textbf{a} and \textbf{b} in the Introduction and helps us understand why a discriminator based feature extractor can achieve good performance in downstream tasks. So the answer is possibility \textbf{b}, in which the features extracted for those samples are allowed to be very different in a meaningful way.

\section{Analyze the counterpart of the discriminator extractor -- the classifier extractor}

As mentioned in the Introduction, to fully understand the discriminator as feature extractor, we should understand its counterpart, the classifier extractor.


Like in the previous section, we start by analyzing the objective function of the classification task, which involve cross-entropy, and can be represented by

\begin{dmath}
    \min_q V(q) = -\frac{1}{N} \sum_{j=1}^{N} \sum_{i=1}^{K} p_i(x_j) \log [q_i(x_j)]
    \label{eq10}
\end{dmath}

In this equation, $N$ is the number of samples; $K$ is the number of classes; $p_i(x_j)$ is the probability that the $j_{th}$ sample belongs to the $i_{th}$ class, where $p_i(x_j) = 1$ if $x_j$ belongs to the $i_{th}$ class, and $p_i(x_j) = 0$ if not; and $q_i(x_j)$ is the probability that that sample belongs to that class judged by the classifier.

Similar to a discriminator, a classifier can also be decomposed to three components as shown in the following equations

\begin{dmath}
    q_i(x_j)=\sigma(y_j)_i
    \label{eq11}
\end{dmath}

\begin{dmath}
    y_j=Af(x_j) + b
    \label{eq12}
\end{dmath}

The first component is the feature extractor $f$; the second component is the affine transformation represented by matrix $A$ and bias vector $b$; and the third component is a softmax function $\sigma$ linking the output of the second component to the class probabilities. The law ruling the output of the second component, $y$, is revealed by the following Proposition.

\textbf{Proposition 6.} For data in different classes in the classification task, the outputs of the affine transformation of the classifier are diverse. For data in the same class $i$, the outputs will be $y_i + \lambda (x_j)\mathds{1}$, where $y_i$ is a vector invariant to $x_j$ in that class $i$, $\mathds{1}$ is the vector with all elements to be $1$, and $\lambda(x_j)$ is from the real number.

\textit{Proof:} In Equation (\ref{eq10}), $\sum_{i=1}^{K} p_i(x_j) \log q_i(x_j)$ equals the cross entropy for the distributions $p$ and $q$, which can be rewritten as $H(p \Vert q) = H(p) + D_{KL}(p \Vert q)$, where $H(p)$ is the entropy of $p$, and $D_{KL}(p \Vert q)$ is the KL divergence \cite{kullback_information_1951} between $p$ and $q$. This will be minimum when $q_i(x_j) \rightarrow p_i(x_j)$, that is $q_i(x_j) \rightarrow 1$ if $x_j$ belongs to the $i_{th}$ class, and $p_i(x_j) \rightarrow 0$ if not. When this happens, considering that the softmax function $\sigma$ is continuous and

\begin{dmath}
    \sigma(z + c) = \sigma(z)
    \label{eq13}
\end{dmath}

\noindent if \cite{goodfellow_deep_2016} and only if (Appendix 1) $c= \lambda \mathds{1}$, the inputs of the $\sigma$, which are the outputs of the affine transformation, will be diverse for samples in different classes, and will be $Af(x_j) + b \approx y_i + \lambda (x_j)\mathds{1}$ for samples in the same class $i$.

Combining this with Equation (\ref{eq12}), we have the following two Theorems controlling the features extracted by a classifier.

\textbf{Theorem 2.} Features extracted by a classifier are guided to be diverse for different classes. Besides this, there may be extra discriminator-style divergences between different samples.

\textit{Proof:} The outputs of the affine transformation are diverse for different class samples (Proposition 6). This means $z_j=Af(x_j) = y_j - b$, which is a vector in the column space of the matrix $A$, is different too. Letting $f(x_j) = f_r(x_j) + f_n(x_j)$, where $f_r$ corresponding to the vector in $A$'s row space and $f_n$ corresponding to the vector in $A$'s nullspace (Proposition 4), $x_j$ in different classes will have different $f_r(x_j)$ (Proposition 5). In addition, $f_n$ is orthogonal to $f_r$, so we can conclude that $f(x_j)$ is diverse for different classes, which means features extracted by a classifier are guided to be diverse for different classes. In addition, for the same reason as shown in the proof of the Theorem 1, extra discriminator-style divergences may exist between different samples, even if they are in the same class.

\textbf{Theorem 3.} If a set of samples are from the same class in the source classification task, their similarity will be partially kept in the features extracted.

\textit{Proof:} In getting samples ${x_j}$s from the same class $i$ in the source classification task, it is proven in Proposition 6 that their affine-transformation outputs will be $Af(x_j) + b \approx y_i + \lambda (x_j)\mathds{1}$. In the case that $\lambda$ does not change with $x_j$, $Af(x_j) \approx y_i - b + \lambda(x_j) \mathds{1}$ will be a constant, thus $f(x_j)$ can be decomposed to
$f_r + f_n(x_j)$ based on Proposition 4 and Proposition 5. In the case that $\lambda$ changes with $x_j$, supposing $x_{j_1}$ and $x_{j_2}$ are two samples from this class with different $\lambda$, the following equation should be satisfied

\begin{dmath}
    A[f(x_{j_1}) - f(x_{j_2})] \approx [\lambda(x_{j_1}) - \lambda(x_{j_2})] \mathds{1}
    \label{eq14}
\end{dmath}

From this, $\mathds{1}$ is in the column space of $A$ and will have a unique row space vector which can be denoted as $f_{r_{\mathds{1}}}$. With this, combined with that the row space vector for $y_i - b$ can be denoted by $f_r$, $Af(x_j)$ can be decomposed to

\begin{dmath}
    f_r + \lambda(x_j) f_{r_{\mathds{1}}} + f_n(x_j)
    \label{eq15}
\end{dmath}

Therefore, in both the two cases, the features extracted will inherit the similarity from the source classification with the help of the unchanging $f_r$.

In a nutshell, these two Theorems show that a classifier extractor guides the features extracted to be diverse for data in different classes in the source task and partially transfers the similarity between data in the same class in the source task to the target task.


The analyses of the counterpart of the discriminator extractor show that feature extractors always allow the features extracted for different samples to be different in a large subspace, no matter whether it is discriminator based or classifier based. But the classifier extractor focusing on the difference of the target samples will guide samples in different classes to have different features in an understandable way. So the performance of the classifier based feature extractor will be better than that of the discriminator based extractor in most cases, as the source classification task is usually similar to the target tasks. However, in the case that the source classification task is very different from the target task or even hurts the target tasks \cite{liu_adversarial_2017, rosenstein_transfer_2005}, the discriminator extractor will be better. Further, the features extracted by the classifier will not be guided to exlude noise between different classes, so a discriminator extractor always has its place and will not be replaced completely. This finding overcomes our intuition and makes discriminator based feature extraction a potential direction of representation learning.

\section{Experiments}

\begin{figure}[t]
   \begin{center}
      \includegraphics[width=0.8\linewidth]{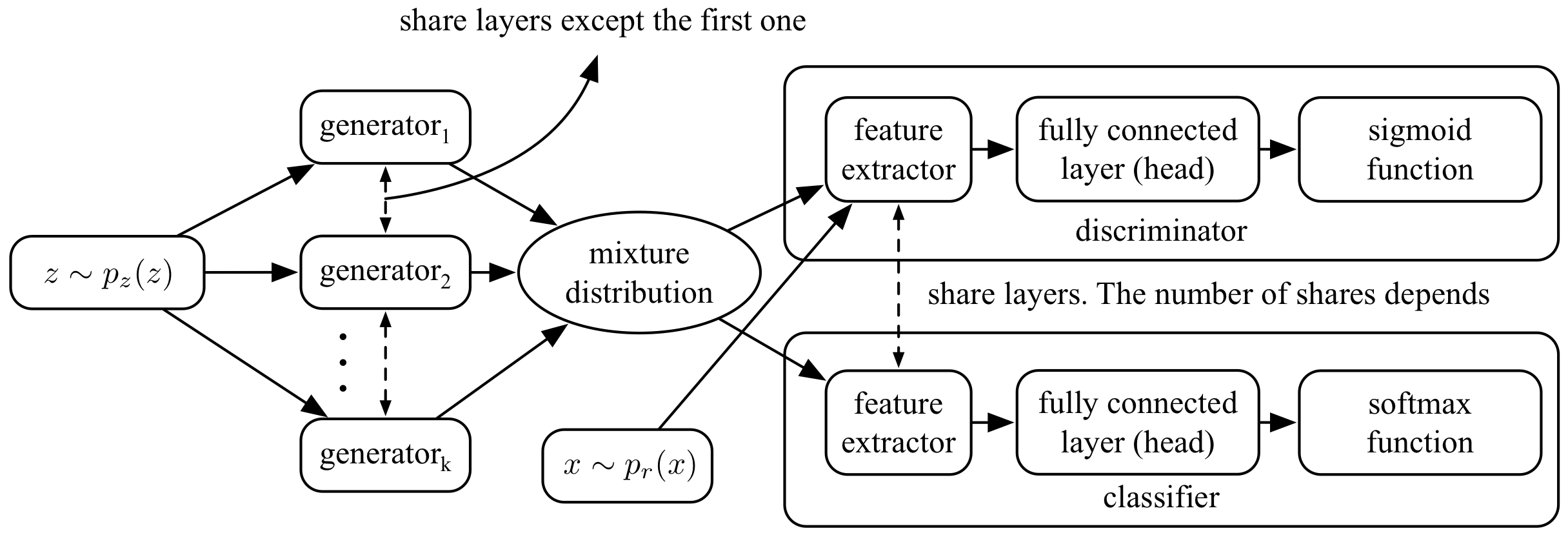}
   \end{center}
      \caption{M-GAN structure. Here we set the mixture distribution to be a multinomial distribution with a uniform probability.}
   \label{fig2}
\end{figure}

In this section, to support our analyses, we compare the performance of classifier and discriminator feature extractors using M-GAN \cite{hoang_multi-generator_2017}. As shown in Figure \ref{fig2}, M-GAN has multiple generators, and it adds a classifier as well as the discriminator. In comparison, traditional GAN \cite{goodfellow_generative_2014} has just one generator and one discriminator. Regarding multiple generators as one generator and using their mixture distribution as the fake data distribution, the discriminator in the M-GAN has a traditional goal to judge whether a sample is from the real data distribution or not. Seeing samples from different generators as from different classes, the classifier here seeks to determine the generator that produces a certain sample. The discriminator and the classifier can share different numbers of neural layers from the last layer to the first layer.

The objective function of this framework is listed in Equation (\ref{eq16}) \cite{hoang_multi-generator_2017}.

\begin{dmath}
    \min_{G_{1:K},C} \max_D V(G_{1:K},C,D) = E_{x \sim p_r(x)} \log D(x) + E_{x \sim p_g(x)}\log[1 - D(x)] - \beta [\sum_{i=1}^K \frac{1}{K} E_{x \sim p_{g_i}} \log C_i(x)]
    \label{eq16}
\end{dmath}

In this Equation, $K$ is the number of generators or classes. $C_i$ is the $i_{th}$ entity in the softmax output of the classifier and represents the probability that sample $x$ is from generator $i$. $\beta$ is greater than $0$ and is a hyperparameter for the tradeoff.

We choose M-GAN in this experiment for three reasons. The first reason is that this is an unsupervised framework with classification as one of its tasks, which helps us compare the discriminator and classifier in the same training settings. The second reason is that the classifier and the discriminator have the same fake samples as inputs, which avoids any difference caused by input data. The third reason is that the number of the shared layers between the discriminator and the classifier can be changed, and can be used to quantitatively measure the influence of the two tasks on the features extracted.

Delving into the third reason, if the two models share all layers in the feature extractor component, the features extracted will be the same for the two models and will be strongly influenced by both the two tasks. If removing the share of the last layer while keeping the shares of the bottom layers of the two extractors, the two extractors will extract different features. When this happens, the feature extracted by the discriminator will more be influenced by the discriminator task but still be attached by the classification task, and vice versa. Continuing this process, features extracted by these two extractors will be more task-specific, and when the two extractors do not share any layer, the features extracted will be influenced only by their own corresponding tasks. These can be used to reflect the feature-extracting capability of different tasks.

The following describes the experimental process: 1. Begin with the discriminator and classifier extractors sharing all layers; 2. train M-GAN for a fixed number of epochs; 3. remove the discriminator and classifier heads (the fully connected layers as shown in Figure \ref{fig2}); 4. add two new heads for the target task to these two extractors respectively; 5. freeze the feature extractors and train the target tasks; 6. compare the improvements on the target tasks from the two extractors; 7. if the current discriminator and classifier extractors share any layers, remove the last share, reset the model weights, and repeat step 2 – 7.

In this experiment, because the classification is the most common and important target task in transfer learning, we choose it as the target task and use the accuracy in the validation set of the target domain as the measure of the target improvements. 

For simplicity, we first use the MNIST \cite{lecun_gradient-based_1998}, which is a set of handwritten digits, as our dataset. With this, we design the model structures and training settings as shown in Table \ref{tb1}. In the experiment process, in step 2 and step 5, we use the full train set for the source unsupervised training and the target supervised learning respectively; and in step 6, we use the full validation set for evaluating the performance of the target classification task. The results are illustrated in Figure 3 and listed in Table \ref{tb2}.

\begin{table}[t]
   \begin{center}
   \resizebox{0.9\linewidth}{!}{ 
   \begin{tabular}{lllllllll}
       \multicolumn{9}{c}{\textbf{Source M-GAN unsupervised training}} \\
       Operation & Kernel & Strides & Padding & Feature Maps & Bias & BN? & Nonlinearity & Shared? \\
       \hline
       $G(z): z \sim N(0, 1)$ & \multicolumn{3}{l}{} & $1 \times 1 \times 128$ & \multicolumn{4}{l}{} \\
       Transposed convolution & $4 \times 4$ & $1 \times 1$ & $0 \times 0$ & $4 \times 4 \times 512$ & no & yes & ReLu & no \\
       Transposed convolution & $4 \times 4$ & $2 \times 2$ & $1 \times 1$ & $8 \times 8 \times 256$ & no & yes & ReLu & yes \\
       Transposed convolution & $4 \times 4$ & $2 \times 2$ & $1 \times 1$ & $16 \times 16 \times 128$ & no & yes & ReLu & yes \\
       Transposed convolution & $4 \times 4$ & $2 \times 2$ & $1 \times 1$ & $32 \times 32 \times 1$ & no & no & Tanh & yes \\
       \hline
       $C(x)$, $D(x)$ & \multicolumn{3}{l}{} & $32 \times 32 \times 1$ & \multicolumn{4}{l}{} \\
       Convolution & $4 \times 4$ & $2 \times 2$ & $1 \times 1$ & $16 \times 16 \times 128$ & no & no & Leaky Rulu (0.2) & depends \\
       Convolution & $4 \times 4$ & $2 \times 2$ & $1 \times 1$ & $8 \times 8 \times 256$ & no & yes & Leaky Rulu (0.2) & depends \\
       Convolution & $4 \times 4$ & $2 \times 2$ & $1 \times 1$ & $4 \times 4 \times 512$ & no & yes & Leaky Rulu (0.2) & depends \\
       Convolution & $4 \times 4$ & $1 \times 1$ & $0 \times 0$ & $1 \times 1 \times 128$ & no & no & & depends \\
       Fully connected & \multicolumn{3}{l}{} & $10$/$1$ & no & no & Softmax/Sigmoid & no \\
       \hline
       Number of generators & \multicolumn{8}{l}{10} \\
       $\beta$ & \multicolumn{8}{l}{0.02} \\
       Batch size for real data & \multicolumn{8}{l}{120} \\
       Batch size for each generator & \multicolumn{8}{l}{12} \\
       Number of epochs & \multicolumn{8}{l}{20} \\
       Learning rate & \multicolumn{8}{l}{0.0002} \\
       Optimizer & \multicolumn{8}{l}{Adam($\beta_1 = 0.5$, $\beta_2 = 0.999$). ($G$, $D$, and $C$ use their own optimizers)} \\
       Weight, bias initialization & \multicolumn{8}{l}{$N(\mu=0,\sigma=0.02)$, $0$} \\
       \hline

       \multicolumn{9}{c}{\textbf{Target classification supervised training}} \\
       Operation & freeze? & \multicolumn{2}{l}{} & Feature Maps & bias & BN? & Nonlinearity & Shared? \\
       \hline
       \multicolumn{4}{l}{} & $32 \times 32 \times 1$ & \multicolumn{4}{l}{} \\
       discriminator/classifier feature extractor & yes & \multicolumn{2}{l}{} & $1 \times 1 \times 128$ & \multicolumn{4}{l}{} \\
       Fully connected & \multicolumn{3}{l}{} & 10 & no & no & Softmax & no \\
       \hline
       Batch size for real data & \multicolumn{8}{l}{120} \\
       Number of epochs & \multicolumn{8}{l}{20} \\
       Learning rate & \multicolumn{8}{l}{0.0002} \\
       momentum & \multicolumn{8}{l}{0.9} \\
       Optimizer & \multicolumn{8}{l}{SGD} \\
       \hline
   \end{tabular}
   }
   \end{center}
   \caption{Model structures and training settings used in the experiment}
   \label{tb1}
\end{table}

\begin{figure}[t]
    \begin{center}
    \includegraphics[width=0.9\linewidth]{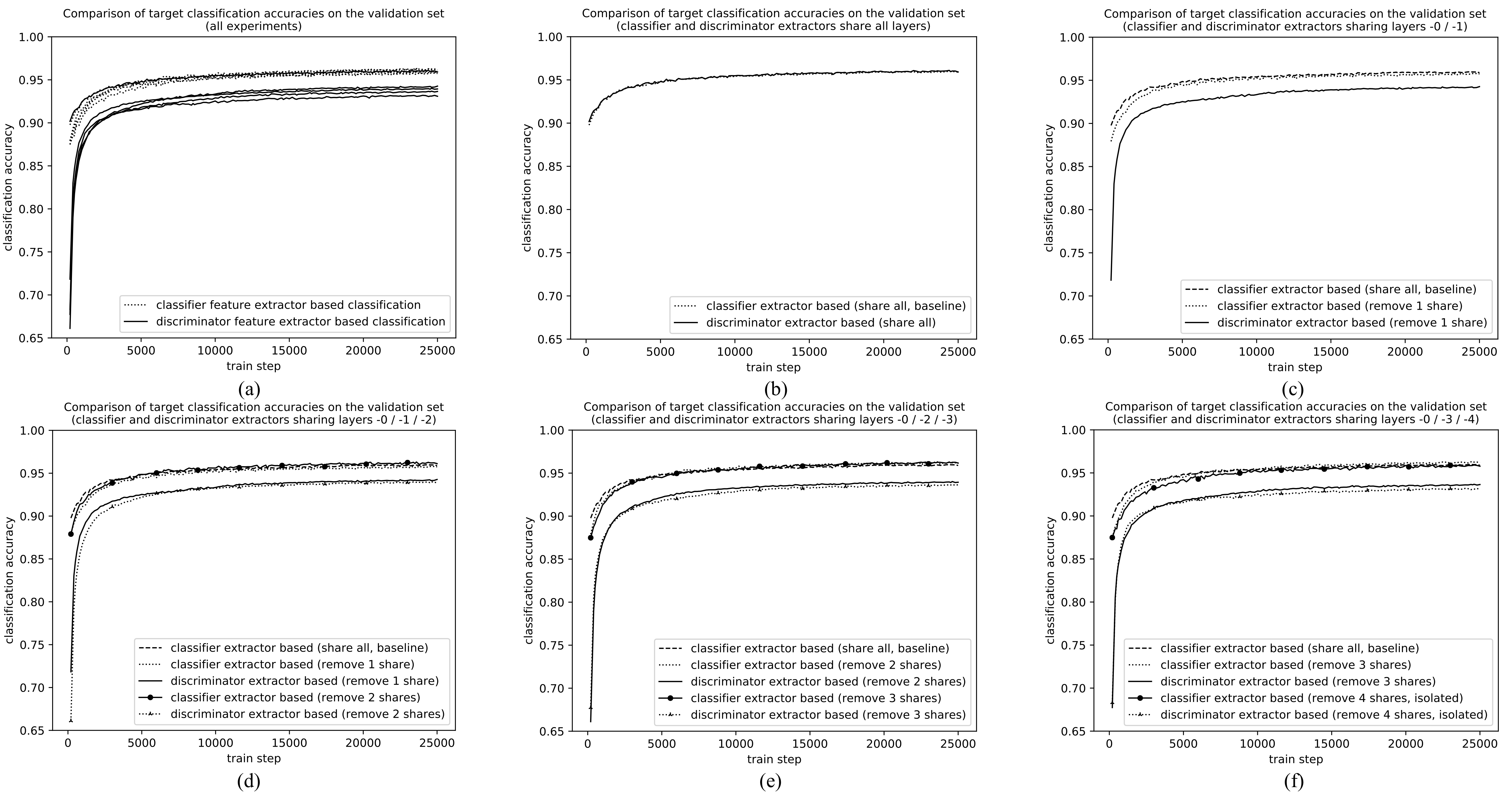}
    \end{center}
    \caption{Target classification accuracies comparison between the discriminator and classifier-based feature extractors. (a) all experiments conducted. (b) the experiment that the two extractors share all layers. The classifier one is set as the baseline. (c) – (d) experiments with different numbers of shared layers between the two extractors.}
    \label{fig3}
\end{figure}

\begin{table}[t]
    \begin{center}
    \resizebox{0.9\linewidth}{!}{ 
    \begin{tabular}{l|lll}
        shares removed & discriminator-extractor-based accuracy  & classifier-extractor-based accuracy  & accuracy difference (classifier - discriminator) \\
        \hline
        0 & 0.9592 & 0.9589 & -0.0003 \\
        1 & 0.9427 & 0.9574 & 0.0147 \\
        2 & 0.9394 & 0.9612 & 0.0218 \\
        3 & 0.9365 & 0.9617 & 0.0252 \\
        4 & 0.9309 & 0.9578 & 0.0269
    \end{tabular}
    }
    \end{center}
    \caption{Target task improvement comparison between discriminator and classifier extractors}
    \label{tb2}
\end{table}

Analyzing these results gives the following: a) When the discriminator and the classifier extractors share all the layers, the difference between the accuracies of the target task based on the two extractors is small, where the accuracy of the discriminator-based one is 0.9592 and that of the classifier-based one is 0.9589. Setting the classifier-based one as the baseline for comparison in the following analysis; b) when one share is removed, the classifier-based approach has an accuracy close to the baseline, but the discriminator-based one becomes much poorer than the classifier-based one, leading to a difference in accuracies of 1.47\%  c) when the two shares are removed, the classifier-based accuracy increases to overperform the baseline, while the discriminator-based one decreases, causing the difference to increase to 2.18\%; d) when one more share is cut, the classifier-based one does not change much, while the discriminator based one decreases making the difference increase to 2.52\%; e) when all shares are broken, both accuracies decrease, but their difference continues to increase, to 2.69\%.

Having less shared layers means more influence of tasks on their extracting features, and the widening gap between the target classification accuracies show that the classification task improves the target task more, even if the source classification task is different to the target one. Considering that no matter how many shares are removed, the performances of the classifier-based one are always close to the baseline, so most of the useful information extracted by the discriminator for the target task is likely to be contained in the classifier extracted features.

We also conducted an experiment based on the CIFAR-10 dataset \cite{krizhevsky_learning_2009}, which is a real image dataset and is more complicated than the MNIST dataset. We adopted the structure similar to the one we used in the MNIST dataset but changed the output of the generator and the input of the discriminator/classifier to have 3 channels. In addition, we also adopted the strategy we used in the MNIST here.

In this case, as the dataset is more complicated, removing too many shared layers will increase the number of parameters to be trained and even cause mode collapse, so we just conducted our experiment with 0, 1, and 2 layer links removed. In all these three cases, the maximum accuracy of the classifier based one is 0.5567, and the maximum accuracy of the discriminator based one is 0.5513. The difference of the accuracy between the classifier based one and the discriminator based one changes from -0.17\% to 6.23\% and finally to 6.62\%. Compared with the MNIST one, the difference is much higher, and it increases when removing more layers. In addition, the classifier achieves maximum accuracy when one shared layer is removed.

These facts, that the classifier performs better and better than discriminator when more shares removed and that the discriminator extractor still has a good performance, support our conclusions that the classifier extractor is better than the discriminator extractor in most cases while the discriminator extractor is a reasonable choice. Further, the result that the maximum accuracy is always achieved when the number of shared layers is not minimum suggests the ability of discriminator to avoid noise.

\section{Discussion}

In this study, we propose a linear-algebra approach to analyze the features extracted. We find that the null space features are the features excluded from the influence of the task -- in the discriminator case, it means getting rid of noise. This suggests a potential transfer learning direction.

In this study, we also revealed the ratio that plays an important role in preventing the mode collapse of GAN, but how to use it to make the training of GAN easier is still not clear. In future study, we will start from this ratio and try to find some metrics to help us train the GAN model.



\section{Conclusions}

In this study, we have proven that the discriminator does not guide the features of samples to be different from the perspective of mode collapse for the entire GAN model, extending a similar conclusion achieved by just analyzing the task of the discriminator. This new perspective is more robust because it rules the whole training process of GAN, while the earlier perspective only focuses on the restriction of the final state of the model. By using linear algebra analyses, we also find that the feature and task are not the same thing, that the task requires the target samples to be assigned into one group, but the model allows the target samples to have very different features in a meaningful way. This helps explains why discriminator extractors can be successful. We also show that the classifier extractor is a better choice in most cases, but the ability of the discriminator extractor to exclude noise is irreplaceable. Further, our analyses on GAN training led to a ratio playing an important role in that process, which is valuable in basic GAN research.

\bibliographystyle{unsrt}  
\bibliography{reference}  

\section{Appendix 1}

\textbf{Proposition 7.} For softmax function $\sigma(x)$, $\sigma(z+c)=\sigma(z)$ if and only if $c=\lambda\mathds{1}$, where $\mathds{1}$ is a vector with all elements to be $1$, and $\lambda$ is from the real number.

\textit{Proof:} $c=\lambda\mathds{1}$ has been proved to be the sufficient condition for $\sigma(z+c)=\sigma(z)$ by Goodfellow, Bengio, and Courville (2016), so we just prove that it is also a necessary condition. The softmax function can be written as:

\begin{dmath}[number=17]
    \sigma(x)_i=\frac{e^{x_i}}{\sum_{j}^{n}e^{x_j}}
    \label{eq14}
\end{dmath}

In this equation, $\sigma(x)_i$ is the $i_{th}$ element of the softmax output vector; $x_i$ is the $i_{th}$ element of the input vector $x$; and $n$ is the length of the two vectors. Denoting $\sigma(x)_i$ to be $s_i$, we have

\begin{dmath}[number=18]
    (s_i - 1)e^{x_i}+\sum_{j\neq i}^{n}{s_ie^{x_j}}=0
    \label{eq15}
\end{dmath}

which can be written as

\begin{equation}
    \begin{bmatrix}
        s_1 - 1 & s_1 & \cdots & s_1 \\
        s_2 & s_2 - 1 & \cdots & s_2 \\
        \vdots & \vdots & \ddots & \vdots \\
        s_n & s_n & \cdots & s_n - 1
    \end{bmatrix} \begin{bmatrix}
        e^{x_1} \\
        e^{x_2} \\
        \vdots \\
        e^{x_n}
    \end{bmatrix} = 0
    \tag{19}
    \label{eq19}
\end{equation}

Based on this equation, left multiplying its two sides by the following invertible matrix

\begin{equation}
    \begin{bmatrix}
        \frac{1}{s_1} & & & \\
        & \frac{1}{s_2} & & \\
        & & \ddots & \\
        & & & \frac{1}{s_n}
    \end{bmatrix}
    \tag{20}
    \label{eq20}
\end{equation}

we get

\begin{equation}
    \begin{bmatrix}
        \frac{s_1 - 1}{s_1} & 1 & \cdots & 1 \\
        1 & \frac{s_2 - 1}{s_2} & \cdots & 1 \\
        \vdots & \vdots & \ddots & \vdots \\
        1 & 1 & \cdots & \frac{s_n - 1}{s_n}
    \end{bmatrix} \begin{bmatrix}
        e^{x_1} \\
        e^{x_2} \\
        \vdots \\
        e^{x_n}
    \end{bmatrix} = 0
    \tag{21}
    \label{eq21}
\end{equation}

In this equation, the matrix in its left side can be eliminated to
\begin{equation}
    \begin{bmatrix}
        -\frac{1}{s_1} & 0 & \cdots & 0 & \frac{1}{s_n} \\
        0 & -\frac{1}{s_2} & \cdots & 0 & \frac{1}{s_n} \\
        \vdots & \vdots & \ddots & \vdots & \vdots \\
        0 & 0 & \cdots & -\frac{1}{s_{n-1}} & \frac{1}{s_n} \\
        0 & 0 & \cdots & 0 & 1 - \frac{1}{s_n} + \frac{\sum_1^{n-1}s_i}{s_n}
    \end{bmatrix}
    \tag{22}
    \label{eq22}
\end{equation}

As $\sum_{1}^{n}s_i=1$, we have $1 - \frac{1}{s_n} + \frac{\sum_1^{n-1}s_i}{s_n}=0$, so that the rank of this matrix is $n - 1$, and the matrix has an 1-dimension null space. Therefore, if $\sigma(x)=\sigma(y)$, the vector $\begin{bmatrix} e^{x_1} & e^{x_2} & \cdots & e^{x_n} \end{bmatrix}^T$ will be equal to the vector $\begin{bmatrix} e^{y_1} & e^{y_2} & \cdots & e^{y_n} \end{bmatrix}^T$ times a constant, thus $x = y + \lambda \mathds{1}$, where $\lambda$ is from the real number.

\end{document}